\documentclass{article}

\usepackage{microtype}
\usepackage{graphicx}
\usepackage{subfigure}
\usepackage{amsmath}
\usepackage{booktabs} 

\usepackage{hyperref}

\usepackage[accepted]{icml2019}

\icmltitlerunning{Periodically retraining classifiers with feedback from a team of end users}

\begin{document}
	
	\twocolumn[
	\icmltitle{Some people aren't worth listening to: periodically retraining classifiers with feedback from a team of end users}
	
	
	\begin{icmlauthorlist}
		\icmlauthor{Joshua Lockhart}{jp}
		\icmlauthor{Samuel Assefa}{jp}
		\icmlauthor{Tucker Balch}{jp}
		\icmlauthor{Manuela Veloso}{jp}
	\end{icmlauthorlist}
	
	\icmlaffiliation{jp}{JP Morgan Artificial Intelligence Research}
	
	\icmlcorrespondingauthor{Joshua Lockhart}{joshua.lockhart@jpmchase.com}
	
	\icmlkeywords{Machine Learning, ICML}
	
	\vskip 0.3in
	]
	
	
	
	\printAffiliationsAndNotice{}  
	
	\begin{abstract}
		Document classification is ubiquitous in a business setting, but often the end users of a classifier are engaged in an ongoing \textit{feedback-retrain} loop with the team that maintain it. We consider this feedback-retrain loop from a multi-agent point of view, considering the end users as autonomous agents that provide feedback on the labelled data provided by the classifier. This allows us to examine the effect on the classifier's performance of unreliable end users who provide incorrect feedback. We demonstrate a classifier that can learn which users tend to be unreliable, filtering their feedback out of the loop, thus improving performance in subsequent iterations.

	\end{abstract}
	
	\section{Introduction}
	\label{section:introduction}	
	
Financial institutions must process huge amounts of textual data as part of their daily business and operations. When implemented and maintained to a high standard, document classification systems can increase productivity of teams throughout such an organisation, from the front office to the back, from sales through HR, and everywhere in between.


	 Consider a team of analysts assigned to monitor news articles that may have a bearing on the fate of publicly traded securities. To cope with such a vast stream of information, the analysts may ask a data science team to build a supervised or semi-supervised machine learning classifier to categorize the news articles as they arrive. News articles could be detected as containing information about company earnings releases, announcements about central bank policy changes, or even environmental events that would affect commodities prices. 

Consider also a team of support staff who share an email mailbox. If a member of the business needs the support team to do something, they send an email to the shared mailbox. The request is then assigned to one of the support staff to work on. To aid the team in processing the stream of support requests, a classifier could be deployed to categorise each one as it arrives, \emph{e.g.} an email could be labelled `Tech Failure' if it contains technical details related to I.T. infrastructure, `Wrong Mailbox' if it is off topic, or even `Severe' if the email body mentions lost revenue or contains profanity, \emph{etc}.
    
	
In both of these cases the business teams have identified the need for a purpose built document classifier. A data science team works with them to train and deploy such a classifier, and it is deployed in a production setting in an effort to reduce workload of the business team.

Unfortunately this is rarely the end of the story. No machine learning classifier will be in production for very long before it provides an incorrect classification of a document. This could be for any number of reasons, including but not limited to out of date or insufficient training data, changing business requirements, or even a change in the format of the data to be classified.
	
Inevitably the data science team will be contacted by the business team if the classifier stops performing well. The data science team can use this feedback to improve the classifier, perhaps retraining the model and redeploying. It is this \textit{feedback-retrain} loop that we focus on in this paper: we consider feedback given by the end users of a classifier (\emph{agents} in a \textit{feedback pool}), and investigate schemes for using this agent feedback to retrain the classifier.
	
	
	An immediate issue that arises in this setting is the following: which agents in the feedback pool provide reliable feedback? Is it wise to blindly incorporate all agent feedback, or should we become more selective? It seems reasonable that more experienced agents in the feedback pool will provide better feedback than less experienced agents. For instance, perhaps a subset of the agents in the pool confuse two categories, and consistently swap one for the other when asked for feedback on a particular labelling of a document. Retraining the classifier on documents with labels provided by such an agent will cause issues with performance.
	

	
	




	Explicitly, we consider how a classifier's accuracy can be improved over time by engaging in a feedback loop with a pool of agents. The agents receive labelled documents from the classifier. If an agent thinks their document has been mis-labelled, they provide the classifier with feedback: an alternate label that they believe is correct. These `re-labelled' documents are collected and used to retrain the classifier, in the hope that this new iteration will provide better performance than the current one.
	
	The goal of the classifier is to identify and filter out agents whose feedback degrades rather than improves performance. Note that our motivating scenarios are those in which each discrete document is sent to one agent to work on. Hence, in our setting each document will receive one piece of feedback: a single alternative label provided by an agent in the pool. This complicates matters because the classifier is unable to consolidate multiple pieces of feedback on a single data point.

	\subsection{Related work}
	\cite{hovy2013learning} consider how trustworthy users can be identified from non-expert annotation services, \emph{e.g.} Amazon Turk. They identify what we would refer to as noisy agents by comparing multiple user's responses to a particular task, which is in contrast to the problem we are trying to solve where at most one user will give feedback on a data point they receive. \cite{vaughan2017making} provides a comprehensive survey on how machine learning models can be trained from crowdsourced labelled data. Concrete examples of work on the issues that arise in this setting are \cite{dalvi2013aggregating}, who consider how multiple inputs from users can be aggregated to learn the ground truth of a binary classification task. \cite{zhang2014spectral}, \cite{Zhu2015OnlineC} and \cite{sinha2018fast} all build on the seminal Dawid-Skene algorithm \cite{dawid1979maximum} for aggregating multiple, often differing opinions on a data point into a classification close to some ground truth. 
	
	This existing literature tends to come from the perspective of \textit{obtaining} labels for data that is intended to be used as training set for a classifier. The problem we consider is an online version of this problem, where instead of relying on a crowd of unknown participants (\emph{c.f. Amazon Turk}) to give us some semblance of a ground truth, we are interested in taking labels from a team of end users with a vested interest in the classifier performing well. In this setting it is appropriate to keep track of the performance of individual users, with the aim of giving them direct feedback on how helpful their feedback is for the classifier. Furthermore, we are interested in the case where there is only one piece of feedback on a labelled item, from a single end user, meaning we are not able to make use of any sort of majority vote based algorithm. 
	
	\section{Approach}
	\subsection{Training and feedback loop}
	We consider an iterative learning scenario over a dataset $X,Y$ partitioned into $N$ equal chunks $(X^{(1)},Y^{(1)}),\dots, (X^{(N)},Y^{(N)})$. Each data point $x\in X$
	has a \textit{true label} $y$ that belongs to the fixed set of target labels $\mathcal{L}$. There is assumed to be a fixed pool of \emph{agents} $A=\{A_1,\dots, A_M\}$.
	Upon receiving a label $l\in \mathcal{L}$, an agent $A_i\in A$ will respond with another label $l'\in \mathcal{L}$, which we refer to as that agent's \textit{feedback}.
	 Note that for the purposes of our simulation, agents simply act as functions $A_i:\mathcal{L}\rightarrow\mathcal{L}$. To elicit `feedback' on a classification, we as the experimenter will simply pass the agent the ground truth for the data point under consideration and use their response to retrain the classifier. 
	  As we will define in Section \ref{sec:agents}, agents can be reliable (act as identity function), noisy (respond with random selection from $\mathcal{L}$), or confused (act as non-identity function on $\mathcal{L}$). 
	
	The iterative learning scenario for a classifier $C_\theta$ over a partitioned data set $(X^{(1)},Y^{(1)}),\dots,(X^{(N)},Y^{(N)})$ with agent feedback pool $A=\{A_1,\dots,A_M\}$ is as follows
	
	\begin{enumerate}
		\item \label{step1} Receive $X^{(i)}=x_1^{(i)},\dots,x_k^{(i)}$. Provide labels $L=\{C(x^{(i)}_1),\dots, C(x^{(i)}_k)\}$.
		\item \label{step2} Receive agent feedback $A(X^{(i)}) := \left(j_1,A_{j_1}\left(y_1^{(i)}\right)\right),\dots,\left(j_k,A_{j_k}\left(y_k^{(i)}\right)\right)$, where each $j_u$ is selected uniformly at random from $\{1,\dots,M\}$.
		\item Based on agent feedback $A(X^{(i)})$ and classifier labels $L$, adjust classifier parameters $\theta$. 
		\item Set $i=i+1$. Go to step \ref{step1}.
	\end{enumerate}
	To elucidate Step \ref{step2}: for each datapoint $x_u$ we randomly select a user $A_{j_u}$, recording the index of the user $j_u$ and their response to the ground truth labelling of that datapoint $A_{j_u}(x_u)$. This is the information the training scheme will use to determine if each piece of feedback is worth using in subsequent training stages.
	
	
	At each iteration, the classifier's accuracy is measured by comparing the classifier's labels $x_1^{(i)},\dots,x_k^{(i)}$ with the ground truth labels $y_1^{(i)},\dots, y_k^{(i)}$ by means of the $F_1$ metric with micro averaging \cite{F1Metric}.
	
	Note that the classifier taking part in this learning scenario receives only the data points and the feedback provided by the pool of agents. The ground truth labels $Y$ are provided to the agents only. This allows us to allow us to gain control over the `personality' of an agent by treating them as functions $A:\mathcal{L}\rightarrow \mathcal{L}$. Consider for example a reliable agent that always provides the ground truth. This can be modelled as the identity function. At the other extreme, an agent that acts as permutation on  $\mathcal{L}$ with no fixed points will be completely useless as a source of training examples: it always gives the wrong class labels.

	The main contribution of this work is a means of identifying unreliable agents during the training of the classifier, then disregarding the feedback. We demonstrate an algorithm, \textsc{PruneNB} that acts as a filter over the feedback provided to the classifier at each iteration. We demonstrate that unreliable agents can be identified from their behaviour in previous iterations and filtered out from the feedback loop. We call such algorithms \emph{training schemes}.
	
	\subsection{Modelling agent feedback}
	\label{sec:agents}
	In our experiments the pool of agents that give feedback to the classifier at each step will consist of agents of the following kinds:
	\begin{itemize}
		\item \emph{Reliable Agents}: respond with the true label for any data point they receive; explicitly $A(l)=l$ for all $l\in \mathcal{L}$.
		\item \emph{Noisy Agents}: respond with a label selected uniformly at random from the set of target labels; explicitly $A(l) = l'$ where $l'\in_R \mathcal{L}$, for all $l\in \mathcal{L}$.
		\item \emph{Confused Agents}:
always respond according to some fixed non-identity function $A:\mathcal{L}\rightarrow\mathcal{L}$. Such agents are deterministic, but consistently incorrect about all or some of the classes in $\mathcal{L}$.
	\end{itemize}
	
	
	
	
	\subsection{Algorithms}
	\begin{algorithm}[tb]
		\caption{\textsc{UpdateTrustTable}}
		\label{alg:updateTrust}
		\begin{algorithmic}
			\STATE {\bfseries Input:} Labels $l_1,\dots,l_k$, a list of agent IDs and their feedback labels $\{(j_1,l_1'),\dots,(j_k,l_k')\}$\\
			\FOR{$i=1$ {\bfseries to} $k$}
			\IF{$l_i = l_i'$}
			\STATE Update agent trust score $\mathcal{T}(j_i)$: \textit{the agent has provided another label, and we agree on it}
			\ELSE
			\STATE Update agent trust score $\mathcal{T}(j_i)$: \textit{the agent has provided another label, but we disagree with it}
			\ENDIF
			\ENDFOR
		\end{algorithmic}
	\end{algorithm}
	\begin{algorithm}[tb]
		\caption{\textsc{Prune}}
		\label{alg:prune}
		\begin{algorithmic}
			\STATE {\bfseries Input:} Stream of $N$ chunks of datapoints, $X^{(1)},\dots,X^{(N)}$, access to feedback agent pool $A=\{A_1,\dots, A_M\}$. 
			\FOR{$i=1$ {\bfseries to} $N$}
			\STATE Get labels from classifier $L:=l_1,\dots,l_k=C(x_1^{(i)}),\dots,C(x_k^{(i)})$
			\STATE Get feedback from agent pool
			$A(X^{(i)}) = (j_1,l_1'),\dots,(j_k,l_k')$.
			\STATE Call \textsc{UpdateTrustTable} with arguments $L$ and $A(X^{(i)})$.
			\STATE $R\gets [~]$
			\FOR{$u=1$ {\bfseries to} $k$}
			\IF{$\mathcal{T}(j_u)$ satisfies $\langle\text{\emph{TrustCondition}}\rangle$}
			\STATE Append $(x_u^{(i)},l_u')$ to list $R$.
			\ENDIF
			\ENDFOR
			\STATE Continue training classifier with labelled data points in $R$.
			\ENDFOR
		\end{algorithmic}
	\end{algorithm}
	
	The general framework of our algorithms is laid out in Algorithm \ref{alg:prune}. For each agent in the feedback pool, the algorithm maintains a \emph{trust score}. Explicitly, an agent's trust score is the quantity
	\begin{align*}\mathcal{T}(i):=\frac{\#\text{times my labels have agreed with agent } A_i \text{'s labels}}{\#\text{times agent } A_i \text{ has provided a label}}\end{align*}
	which is updated at each iteration of the feedback loop. In Algorithm \ref{alg:prune} we refer to a generic $\langle\text{\emph{TrustCondition}}\rangle$, which can be one of two conditions
	\begin{itemize}
		\item \textsc{MeanPrune}: if $\mathcal{T}(i) > \sum_{j=1}^M\mathcal{T}(A_j)/M$ then accept, otherwise reject.
		\item \textsc{ThresholdPrune}: for some fixed threshold $c$, if $\mathcal{T}(i) > c$ then accept, otherwise reject.
	\end{itemize}
	
	In the next section we outline the experiments we perform to investigate the performance of these training schemes.
	
	\subsection{Experimental methodology}
	We call the following an $N$-stage \textit{trial} of a classifier $C$ on a set of data points $X$ with ground truth labels $Y$, under feedback from a pool of agents $A$.
	\begin{enumerate}
		\item Shuffle $X,Y$. Partition $X,Y$ into $N$ chunks $(X^{(1)},Y^{(1)}),\dots, (X^{(N)},Y^{(N)})$.
		\item Pre-train classifier $C$ on $X^{(1)}$ and $Y^{(1)}$.
		\item Run \textsc{Prune} implementation on $X^{(2)},\dots, X^{(N)}$ with classifier $C$ and feedback pool $A$. Record $F_1$ score at each iteration $i$ of the outer loop by comparing the classifier's labels $L$ with the true labels $Y^{(i)}$.
	\end{enumerate}
	
	The classifier we use as $C$ is a Multinomial Naive Bayes classifier. The partial\_fit() function provided in the Scikit-Learn \cite{scikit-learn} implementation of this classifier is used to continue training at each stage of the trial. We henceforth refer to our training schemes as \textsc{MeanPruneNB} and \textsc{ThresholdPruneNB} respectively.
	
	To judge the performance of \textsc{MeanPruneNB} and \textsc{ThresholdPruneNB} we run $100$ repetitions of a $10$-stage trial as described above. We are interested in the $F_1$ score at each stage of the feedback-retraining loop, so we save each $10$-vector of $F_1$ scores after each trial $\vec{t_1},\dots, \vec{t_{100}}$, and take the mean over the $100$ repetitions $\sum_{i=1}^{100}t_i/100$ to obtain an estimate of the performance at each stage.
	
	
	
	\subsubsection{Datasets}
	We run our experiments on two datasets. The first dataset is the 20 Newsgroups Dataset \cite{20NG}, obtained in a `pre-vectorized' form using the Scikit-learn library \cite{scikit-learn}. We select all emails from the `sci.med',`comp.graphics', `talk.politics.mideast', and `sci.space' mailing lists, giving us a dataset with $\{594,584,564,593\}$ data points in the respective classes. The classification task we consider is to take an email from this set, and label it with the newsgroup it is from.
	
	The second data set is taken from the Reuters news document data set \cite{reuters}. Specifically, we use the 90 class subset of the Reuters-21578 data set distributed as part of the nltk library \cite{nltk}. We select the news articles that belong to the categories $\mathcal{C}=\{$`earn', `acq', `money-fx', `grain', `crude', `trade', `interest', `sugar', `corn', `ship'$\}$. Articles in the Reuters dataset can belong to more than one category. To simplify matters we will not consider the problem of multi-label classification, so we remove all articles in this dataset that belong to more than one category in $\mathcal{C}$ (\emph{i.e.} we filter out articles that aren't labelled with at least one category in $\mathcal{C}$. Then we filter out articles in that reduced set that belong to more than one category in $\mathcal{C}$). This leaves us with a dataset with
	$\{2847,
	1609,
	369,
	217,
	315,
	318,
	203,
	104,
	2,
	119\}$ datapoints in each respective category. This dataset is then vectorised using the TfidfVectorizer class in Scikit-learn.
	
	The next section contains details on the benchmarks we use to contextualise the performance of the training schemes we consider.
	
	
	\subsubsection{Trusting and Discerning Classifiers}
	As a benchmark we consider two training schemes that represent different extremes: complete trust of all agents; and maximal discernment as to the nature of each agent.
	
	At each round $i$, the \textsc{TrustingNB} scheme accepts all feedback it receives from the users as truthful and useful, continuing the training of the classifier with the datapoints $x_1^{(i)},\dots,x_k^{(i)}$ and feedback labels it received from the agents $l_1',\dots,l_k'$. This is the lower bound benchmark that any training scheme should beat.
	
	Conversely, the \textsc{DiscerningNB} scheme is imbued with perfect knowledge of the nature of each agent in the pool (reliable, noisy, or confused). This scheme will only train on feedback from agents it knows to be reliable (recall the reliable agents always give the ground truth labelling). Thus, the \textsc{DiscerningNB} scheme represents the best possible way in our setting of training the classifier at each stage of feedback: giving it the most possible ground truth available at each stage.
	\section{Experiments}
		\begin{figure}[ht!]
		\begin{center}
			\centerline{\includegraphics[width=\columnwidth]{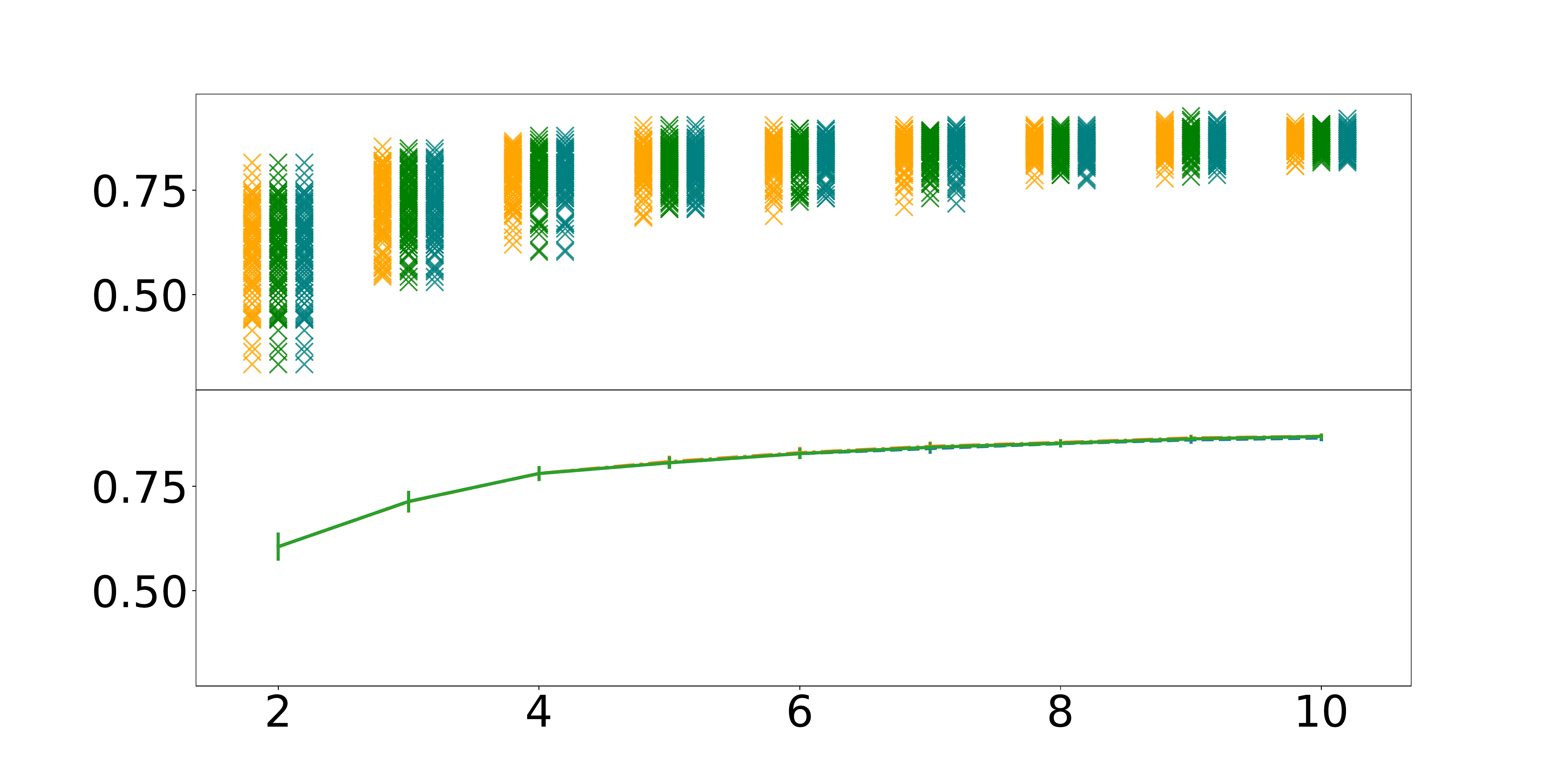}}
			\centerline{\includegraphics[width=\columnwidth]{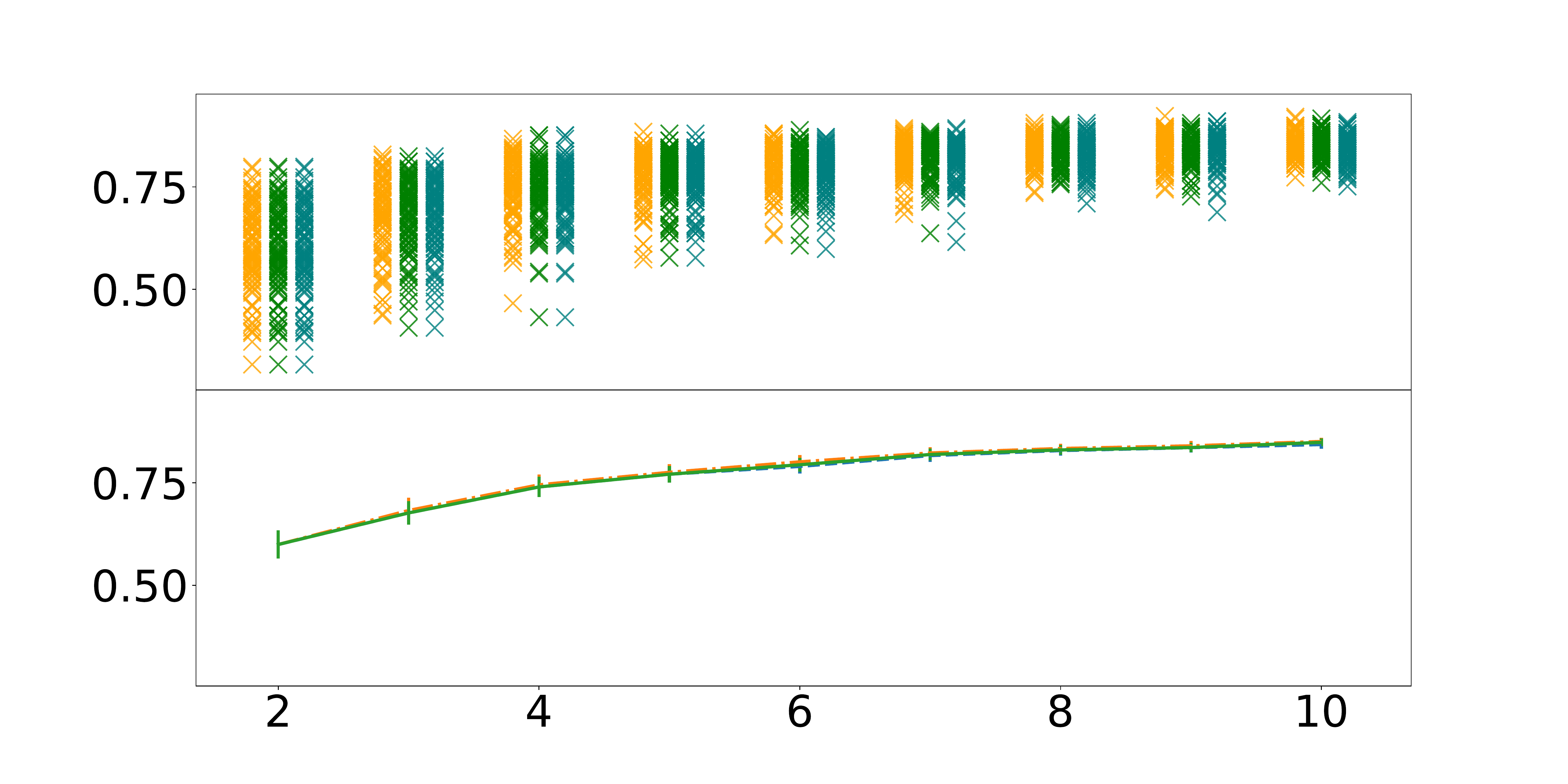}}
			\centerline{\includegraphics[width=\columnwidth]{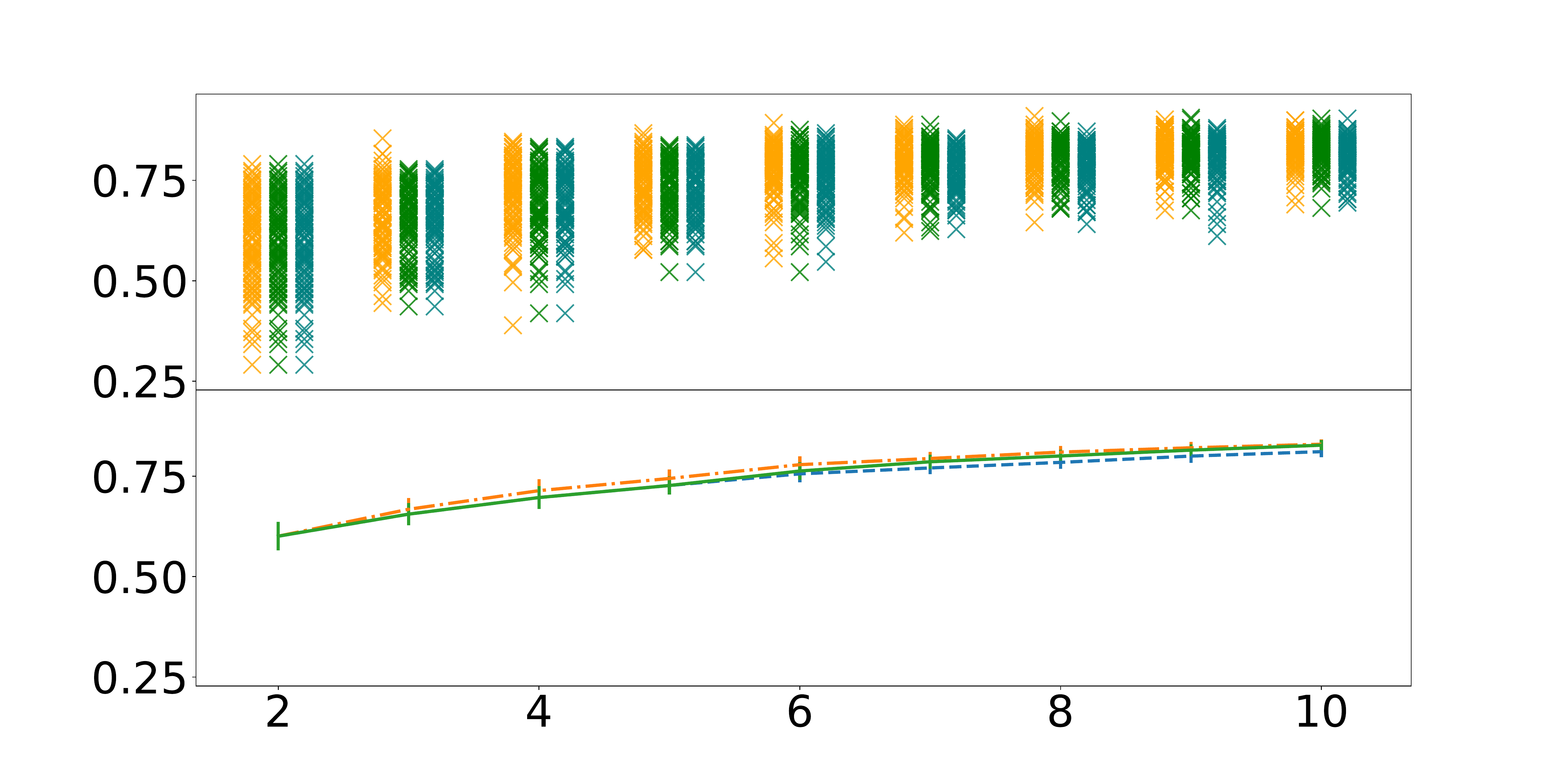}}
			\centerline{\includegraphics[width=\columnwidth]{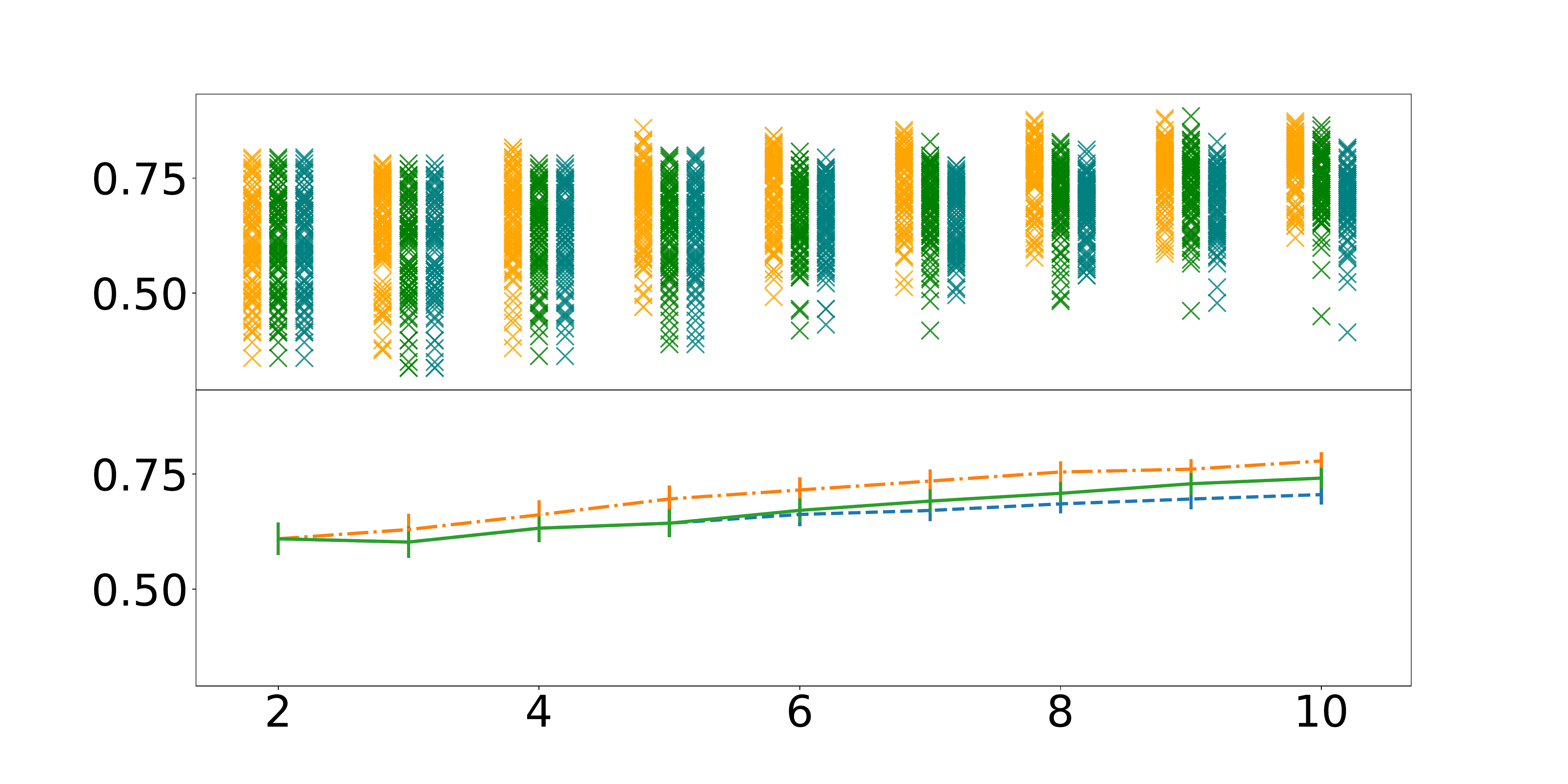}}
			\caption{Performance of \textsc{DiscerningNB} (orange, left/dot+dash), \textsc{ThresholdPruningNB} (green, middle/solid), and \textsc{TrustingNB} (teal, right/dash) on `drip fed' data from Newsgroups with noisy feedback from a pool of five agents. Error bars are standard error, sample size 100. X axis denotes the stage of the trial ($i$), Y axis is $F_1$ score of classifier at that stage. From top to bottom: four reliable agents, one noisy agent; three reliable agents, two noisy agents; two reliable agents, three noisy agents; one reliable agents, four noisy agents.}
			\label{noisy_ng}
		\end{center}
	\end{figure}
	
	\begin{figure}[ht!]
		\begin{center}
			\centerline{\includegraphics[width=\columnwidth]{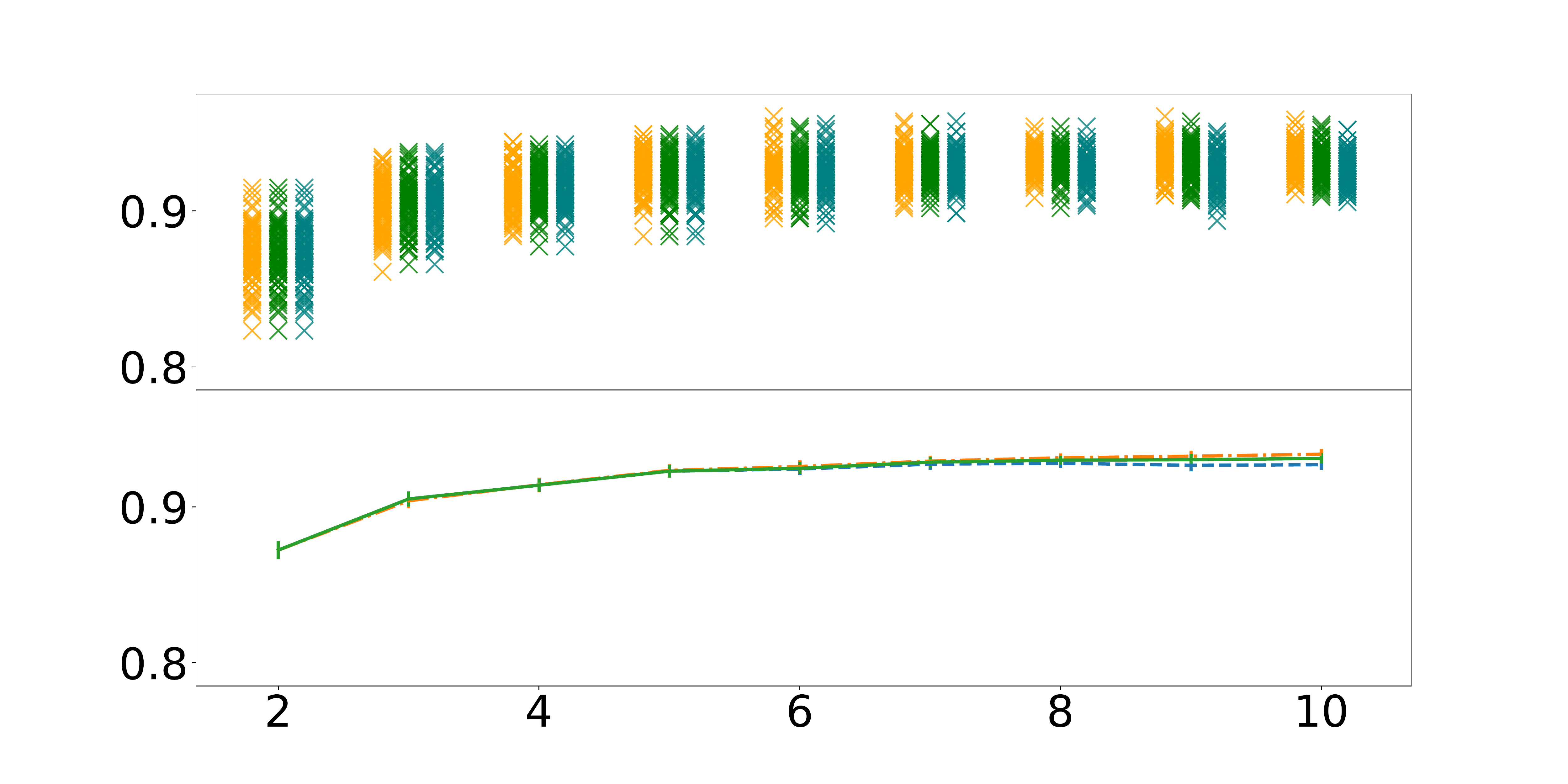}}
			\centerline{\includegraphics[width=\columnwidth]{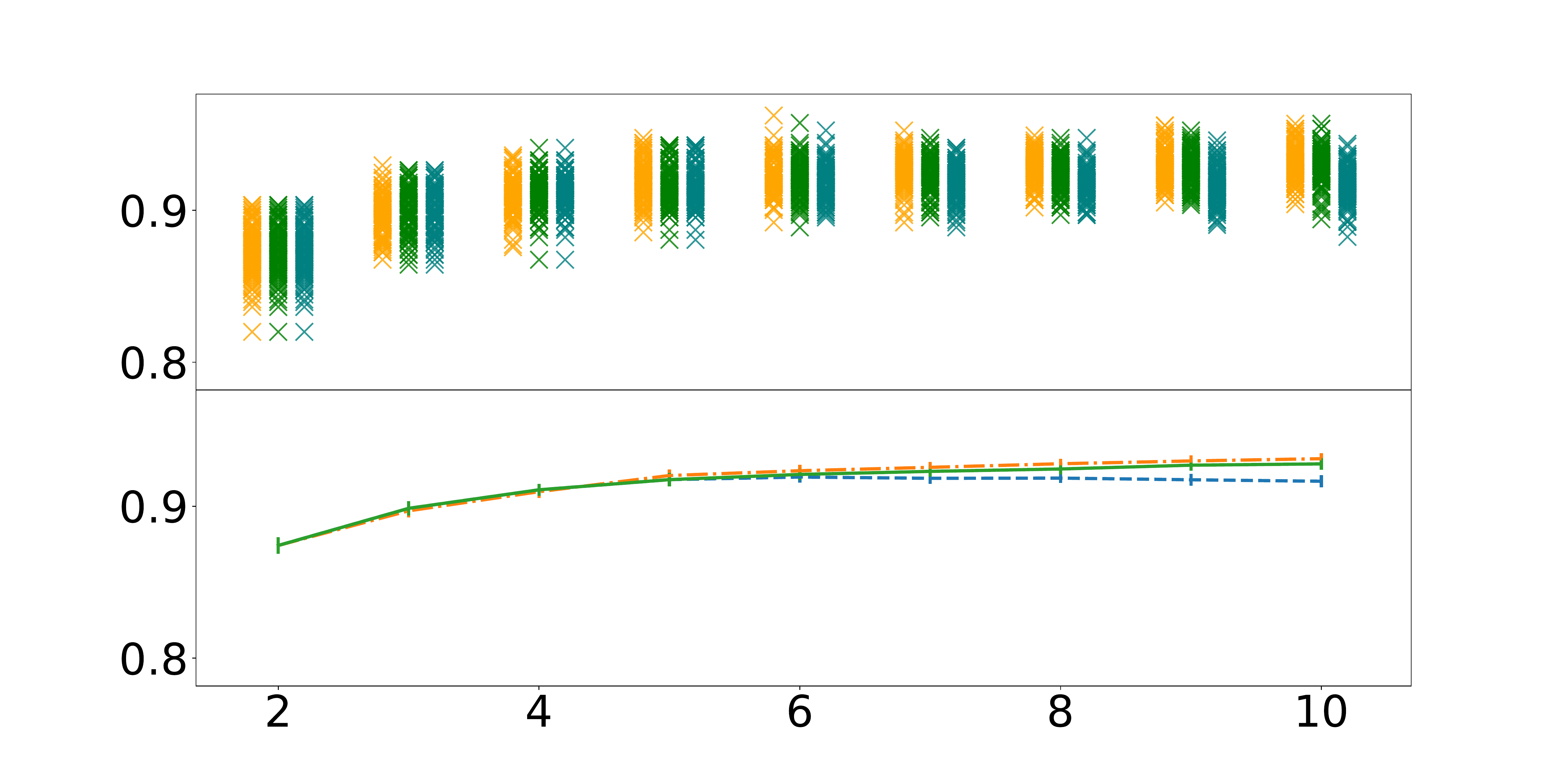}}
			\centerline{\includegraphics[width=\columnwidth]{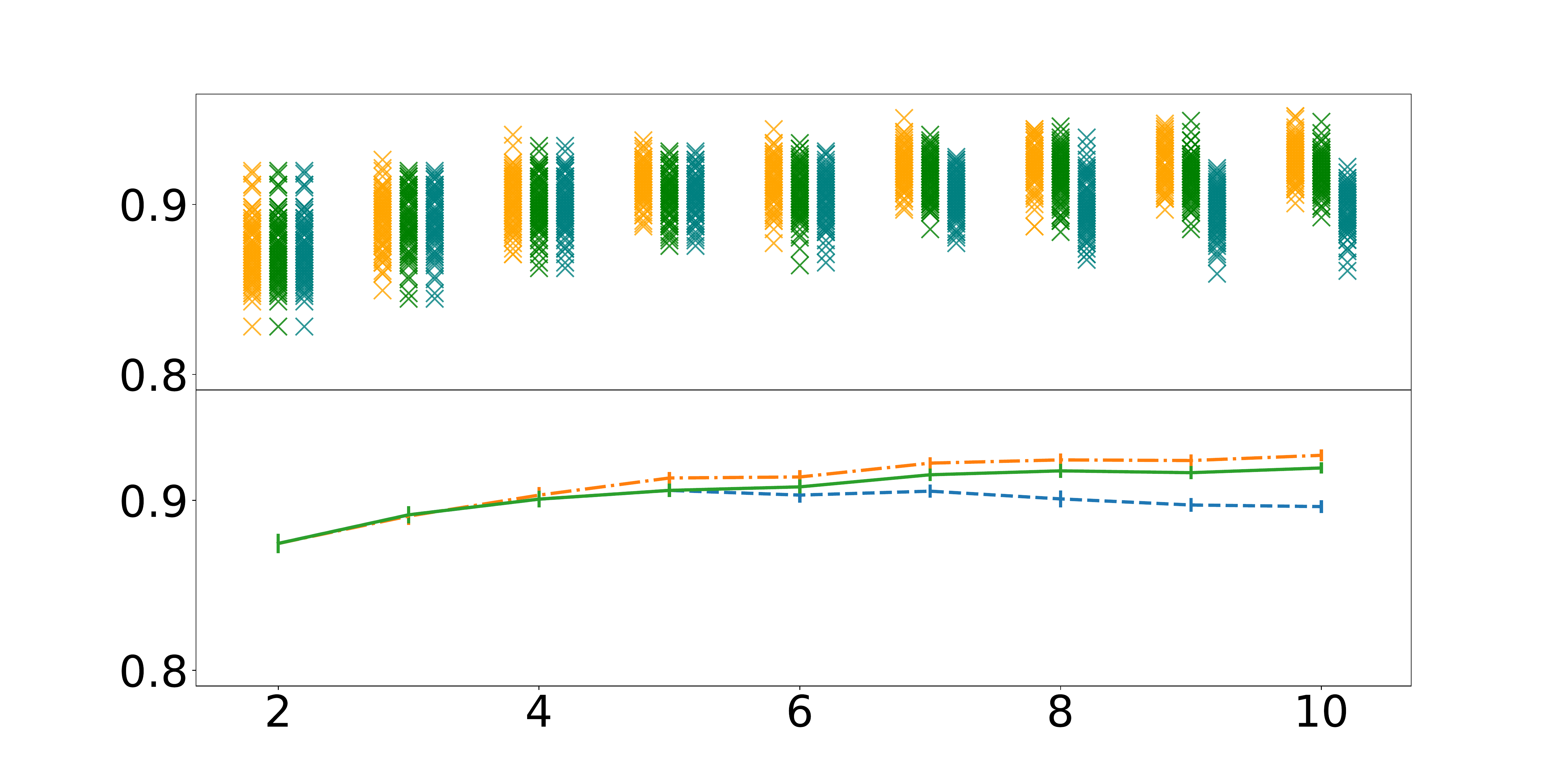}}
			\centerline{\includegraphics[width=\columnwidth]{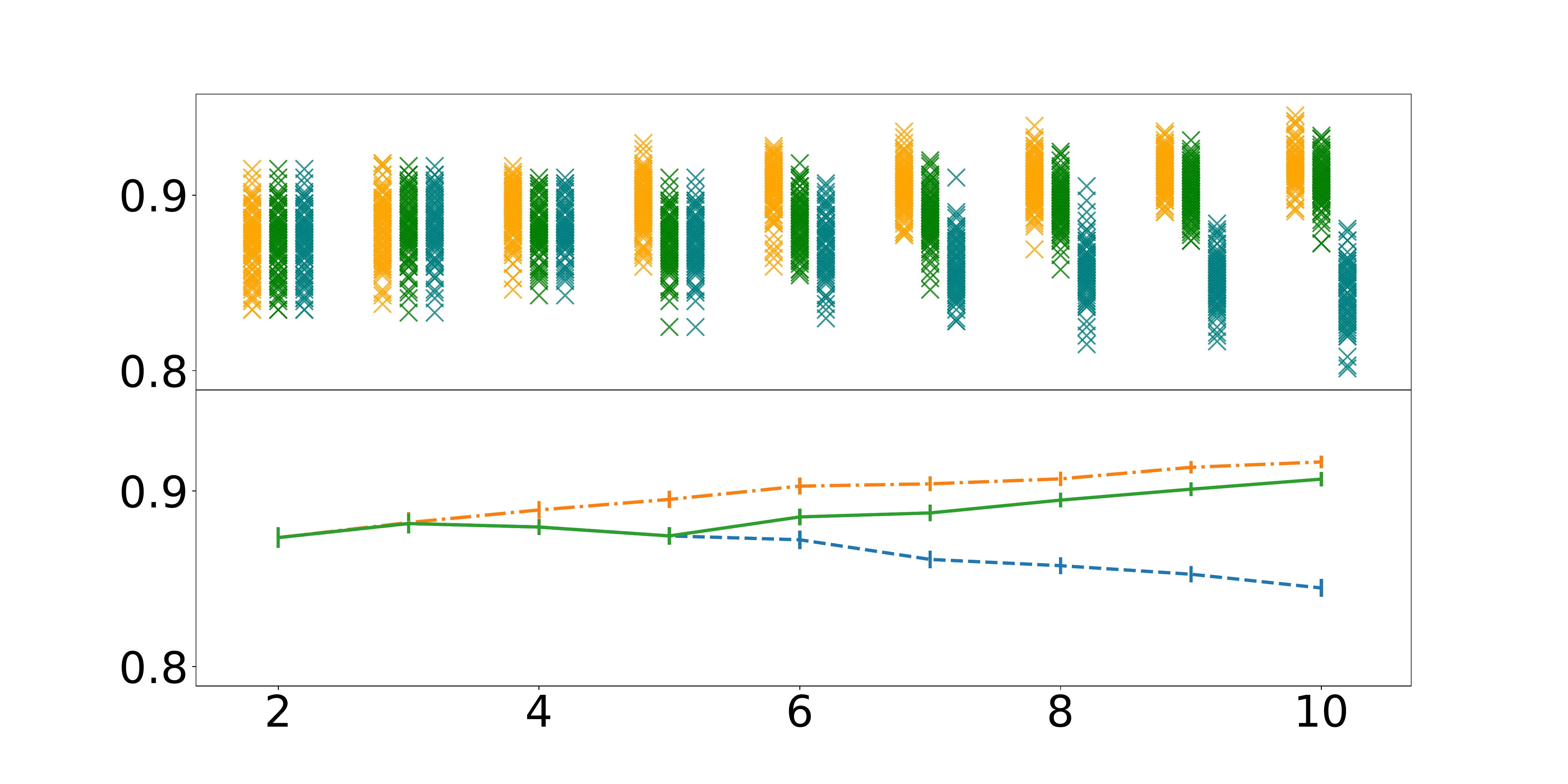}}
			\caption{Performance of \textsc{DiscerningNB} (orange, left/dot+dash), \textsc{ThresholdPruningNB} (green, middle/solid), and \textsc{TrustingNB} (teal, right/dash) on `drip fed' data from Reuters data set with noisy feedback from a pool of five agents. Error bars are standard error, sample size 100. X axis denotes the stage of the trial ($i$), Y axis is $F_1$ score of classifier at that stage. From top to bottom: four reliable agents, one noisy agent; three reliable agents, two noisy agents; two reliable agents, three noisy agents; one reliable agent, four noisy agents.}
			\label{noisy_r}
		\end{center}
	\end{figure}
	\subsection{Learning in the presence of noisy agents}
First, we consider the case where the feedback pool is made up of reliable agents and noisy agents. In Figures \ref{noisy_ng} and \ref{noisy_r} we compare the performance of \textsc{TrustingNB}, \textsc{DiscerningNB}, and our algorithm \textsc{ThresholdPruneNB} on a pool of five agents, with different proportions of noisy users.

During the research we found that performance improved if we made the \textsc{PruneNB} schemes trust all agents until after stage 3 of the feedback-retrain loop. We call this the `burn-in' period of the scheme. During the burn-in period we do no filtering of agents, and the scheme acts like \textsc{TrustingNB}. After the burn-in period is over, the scheme filters agents according to its $\langle\text{\textit{TrustCondition}}\rangle$.
	
	The classifiers were tested simultaneously, each receiving the same chunk of data and agent feedback at each stage of the training. 
	For each experiment we provide two plots, a lower plot which shows a line with error bars and an upper plot which shows data points marked with an `x'. The shared X axis of the upper and lower plots contains the stage $i$ of the feedback loop. The $Y$ axis is the $F_1$ score at each stage. The lower plot contains the mean of this number over the 100 repetitions of the trial, with standard error bars. The upper plot attempts to provide a visualisation of the distribution of this number for each of the three classifiers at each stage. Each `x' in the upper plot corresponds to the $F_1$ score obtained at each stage of one of the 100 trials. Each X position in the upper plot is split into three, one for each classifier. From left to right these are the \textsc{DiscerningNB}, \textsc{ThresholdPruneNB}, and \textsc{TrustingNB} classifiers. The \textsc{ThresholdPruneNB} classifier was run with trust score threshold of $0.3$ and a burn-in of $3$ steps.
	
	
	
	\subsection{Learning in the presence of confused agents}
	Recall that a confused agent is one that answers according to some non-identity function on the labels representing a mistaken belief. We now consider the case where the feedback pool is made up of four reliable agents and a single confused agent. In the experiment on the Newsgroups dataset, 
	the confused agent maps the labels `sci.med' and `comp.graphics' to `comp.graphics', and the labels `talk.politics.mideast' and `sci.space' to `talk.politics.mideast'. In the experiment on the Reuters dataset, the agent will map the labels  `earn', `acq', `money-fx', `trade', and `interest' to the single label `money-fx', and all other labels to `crude'.
	
	We compare the performance on Newsgroups and Reuters datasets of \textsc{TrustingNB}, \textsc{DiscerningNB}, and our algorithm \textsc{MeanPruneNB} on a pool of four reliable agents and the confused agents described above. We plot the results for the Newsgroups dataset in Figure \ref{confused_ng}, and the results for the Reuters dataset in \ref{confused_r}.
	
	Again, schemes were tested simultaneously, each receiving the same chunk of data and agent feedback labels at each stage of the training. There was no burn in period used for these experiments. The lower plot shows the mean and standard error of the $F_1$ score at each stage of the trial, the upper plot shows a plot of the distribution $F_1$ scores at each stage for each classifier.
	
	\section{Discussion of results}
	\subsection{Effectiveness in the presence of noisy agents}
	Figures \ref{noisy_ng} and \ref{noisy_r} concern the performance of the \textsc{ThresholdPruneNB} classifier in the presence of noisy agents. Remarkably, \textsc{TrustingNB} is relatively resilient to noisy feedback from a small subset of the feedback pool, consider in particular the performance illustrated in the top plot of Figure \ref{noisy_ng}: a single noisy agent has barely any effect on the $F_1$ score, almost matching the performance of the \textsc{DiscerningNB}.
	
	The impact of noise becomes more apparent in the case where the majority of agents are providing noisy labels to the classifier, see the case where we have a pool consisting of a single reliable agent and four noisy agents. The impact of noise is most apparent in the Reuters dataset, illustrated in bottom plot of Figure \ref{noisy_r}, where we see the \textsc{TrustingNB} lower benchmark classifier substantially underperform the \textsc{DiscerningNB} upper benchmark classifier. 
	
	Notice the overall performance impact of noisy agent feedback which manifests itself in a wider spread of $F_1$ values. See the widening of the $F_1$ distributions in all stages in the bottom two plots in both Figures \ref{noisy_ng}: while the mean stays relatively similar, the $F_1$ distribution is noticeably wider, meaning that occasionally the classifier performs much worse than usual. We conjecture that this is due to the fact that even in the upper benchmark classifier \textsc{DiscerningNB} case, the classifier trains on substantially less training data, but the $F_1$ score is tested on the full chunk at each stage.
	
	The performance increase afforded by the \textsc{ThresholdPruneNB} scheme over the lower benchmark classifier \textsc{TrustingNB} is very slight in the Newsgroups dataset. The effect manifests itself in the slight increase in stage 8 of the bottom plot in Figure \ref{noisy_ng}, but the \textsc{TrustingNB} value falls within the standard error bar of the \textsc{ThresholdPruneNB} mean. Nevertheless, examining the $F_1$ distribution at stages 7 and 8 we see that the weight of the \textsc{ThresholdPruneNB} distribution (middle, green) seems to fall slightly higher than the \textsc{TrustingNB} distribution (right, teal). The effect is very slight. 
	
	More convincing evidence of the performance increase offered by the \textsc{ThresholdPruneNB} scheme is illustrated in the Reuters dataset. Here we see an unambiguous boost in performance in the one reliable, four noisy agent trial (bottom plot, Figure \ref{noisy_r}) in stages 6, 7 and 8.
	
	
		\begin{figure}[ht!]
		\begin{center}
			\centerline{\includegraphics[width=\columnwidth]{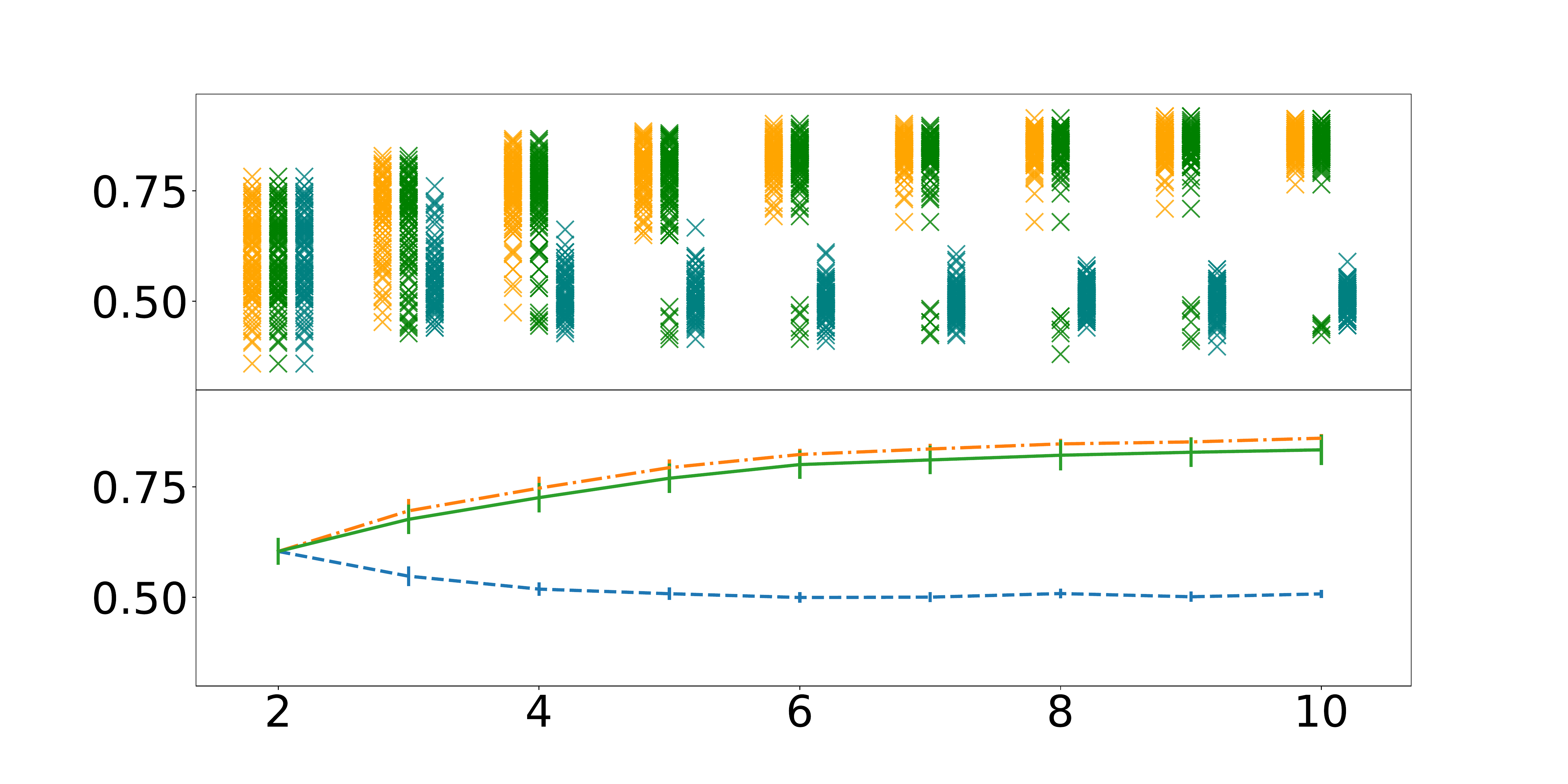}}
			\caption{Performance of \textsc{DiscerningNB} (orange, left/dot+dash), \textsc{ThresholdPruningNB} (green, middle/solid), and \textsc{TrustingNB} (teal, right/dash) on `drip fed' data from the Newsgroups dataset, with a pool of four reliable agents and a single confused agent. X axis denotes the stage of the trial ($i$), Y axis is $F_1$ score of classifier at that stage. Error bars are standard error, sample size 100. }
			\label{confused_ng}
		\end{center}
	\end{figure}
	
	\begin{figure}[ht!]
		\begin{center}
			\centerline{\includegraphics[width=\columnwidth]{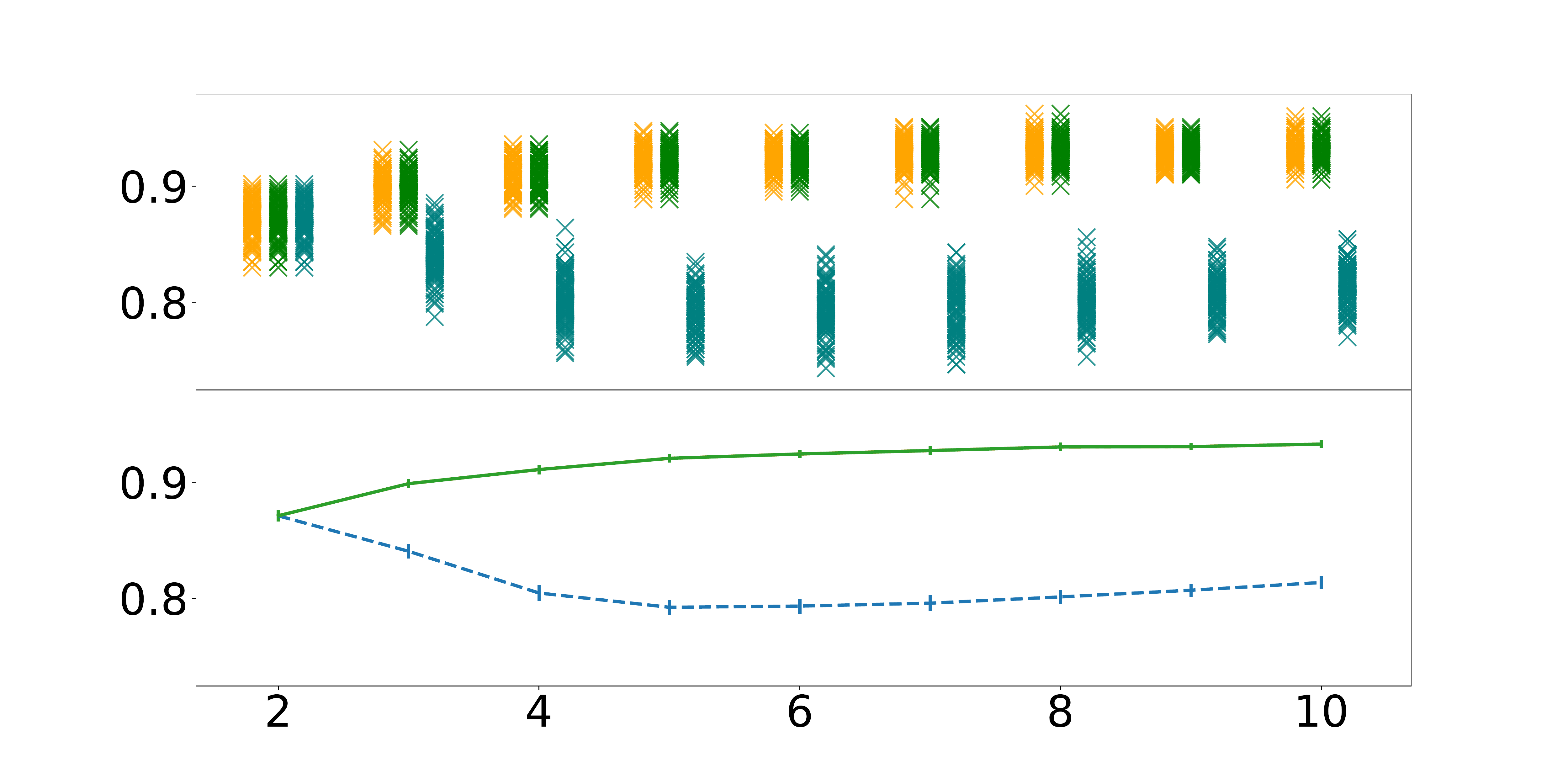}}
			\caption{Performance of \textsc{DiscerningNB} (orange, left/dot+dash), \textsc{ThresholdPruningNB} (green, middle/solid), and \textsc{TrustingNB} (teal, right/dash) on `drip fed' data from the Reuters dataset, with a pool of four reliable agents and a single confused agent. X axis denotes the stage of the trial ($i$), Y axis is $F_1$ score of classifier at that stage. Error bars are standard error, sample size 100. }
			\label{confused_r}
		\end{center}
	\end{figure}

	\subsection{Effectiveness in the presence of confused agents}
	Figures \ref{confused_ng} and \ref{confused_r} illustrate how the three schemes perform on a feedback pool with four reliable users and a confused agent. Unlike a single noisy agent, a single confused agent causes a substantial performance degradation in the \textsc{TrustingNB} benchmark, so much so that we don't consider more than one.
	
	In both datasets we see that 
	our algorithm immediately starts to outperform the lower benchmark \textsc{TrustingNB}. For the Reuters dataset it reaches the same level of performance as the upper benchmark, \textsc{DiscerningNB}. This level of performance is not reached on the Newsgroups dataset. From examining the execution traces of our experiments, we conjecture that this is because in some repetitions of the trial the trust score of some reliable agents can very occasionally dip below the mean, due to errors made by the classifier itself in the early stages of the training. The errors made by the classifier cause it to disagree with the reliable agents, which means they will be taken to have low trust scores. This can compound, and cause the classifier to erroneously distrust one or more of the reliable classifiers throughout all of the subsequent feedback-retrain stages.  
	
Indeed the \textsc{MeanPruneNB} very occasionally performs worse than the \textsc{TrustingNB} scheme. This manifests itself in the wide spread in the $F_1$ distribution in the later stages, an effect most apparent in Figure \ref{confused_ng}. In stage 8 of this figure we note that the distribution has become bimodal. Subsequent work should address this issue, which we feel has arisen due to the `misplaced trust' issue we discuss above.
	
	
	\section{Conclusion}
	We have considered a problem that faces any data science team that deploys a classifier in a production setting: just how do we retrain it when it (inevitably) provides the \emph{users} of that classifier with labels they disagree with? The natural way of achieving this is to ask the users themselves to provide feedback on the outputs of the classifier. Either they agree with the labelling, in which case they send nothing back and the classifier takes this as an implicit confirmation of that label, or they disagree and choose the label from the list they consider more appropriate. An inevitable issue that will arise is that not all of the end users of a classifier will provide meaningful feedback: perhaps the user is disengaged with the process, misinterpets the meaning of some of the target labels, or even misunderstands what the feedback mechanism means. We have attempted to model these feedback scenarios in a multi-agent learning setting: in our imaginary scenario we have posited that there is a team of users (agents) who must process a set of documents according to some business need. These documents are labelled by the classifier, and distributed amongst the agents to work on. If an agent wants to provide feedback on one of the documents they have been given, they give this to the classifier. These feedback `relabellings' are collected and used to periodically retrain the classifier.
	This multi-agent setting provides a compelling new consideration not present in the crowdsourced training data literature: if an agent keeps making mistakes then we'd like to identify them. Perhaps that agent is a new hire and thinks that one label means something else. A system that could identify their confusion could flag this, and the agent could be gently provided with documentation on what the labels mean. Alternatively, agents who provide the system with random feedback labels when they feel like it could be identified and suitable action could be taken.
	
	We have presented a scheme for training a classifier in this iterative setting, the \textsc{Prune} wrapper around the Multinomial Naive Bayes classifier. We have shown how agents in the feedback pool can be assigned a \textit{trust score}, that can be used to segment agents into those who give feedback we want to listen to (high trust score agents) and those whose feedback we have grown to distrust (low trust score agents). We have considered two different ways of segmenting users based on these scores: an explicit threshold cut off (\textsc{ThresholdPruneNB}) and an arithmetic mean cut off (\textsc{MeanPruneNB}). We have demonstrated that these two schemes can cope with noisy users, and confused users respectively.
	\section*{Acknowledgements} J.L. thanks Andy Alexander, Ayham Alajdad, Angelos Filos, Mahmoud Mahfouz for their ongoing input to this work, and Greg Sidier for helpful discussions.
	\section*{Disclaimer} This paper was prepared for information purposes by the Artificial Intelligence Research group of JPMorgan Chase \& Co and its affiliates (“JP Morgan”), and is not a product of the Research Department of JP Morgan.  JP Morgan makes no representation and warranty whatsoever and disclaims all liability, for the completeness, accuracy or reliability of the information contained herein.  This document is not intended as investment research or investment advice, or a recommendation, offer or solicitation for the purchase or sale of any security, financial instrument, financial product or service, or to be used in any way for evaluating the merits of participating in any transaction, and shall not constitute a solicitation under any jurisdiction or to any person, if such solicitation under such jurisdiction or to such person would be unlawful.   
	
	\textcopyright~2020 JPMorgan Chase \& Co. All rights reserved.
	
	\bibliography{main}
	\bibliographystyle{icml2019}
\end{document}